\documentclass[letterpaper]{article} 
\usepackage[submission]{aaai25}  
\usepackage{times}  
\usepackage{helvet}  
\usepackage{courier}  
\usepackage[hyphens]{url}  
\usepackage{graphicx} 
\usepackage{fancyvrb} 
\usepackage{fvextra}  
\usepackage{tcolorbox}

\usepackage{natbib}  
\usepackage{caption} 
\usepackage{algorithm}
\usepackage{algorithmic}
\usepackage{booktabs}
\usepackage{amssymb}
\usepackage{xcolor}

\usepackage{newfloat}
\usepackage{listings}
\DeclareCaptionStyle{ruled}{labelfont=normalfont,labelsep=colon,strut=off} 
\floatstyle{ruled}
\newfloat{listing}{tb}{lst}{}
\floatname{listing}{Listing}

\title{Beyond English: Evaluating Automated Measurement of Moral Foundations in Non-English Discourse with a Chinese Case Study}
\author{Calvin Yixiang Cheng\textsuperscript{\rm 1}, 
    Scott A. Hale\textsuperscript{\rm 1}
    }

\affiliations{
    \textsuperscript{\rm 1} Oxford Internet Institute, University of Oxford\\
    {calvin.cheng}@oii.ox.ac.uk, {scott.hale}@oii.ox.ac.uk

}

\begin{document}

\maketitle

\begin{abstract}
This study explores computational approaches for measuring moral foundations (MFs) in non-English corpora. Since most resources are developed primarily for English, cross-linguistic applications of moral foundation theory remain limited. Using Chinese as a case study, this paper evaluates the effectiveness of applying English resources to machine translated text, local language lexicons, multilingual encoder-only language models, and decoder-only large language models (LLMs) in measuring MFs in non-English texts. The results indicate that machine translation and local lexicon approaches are insufficient for complex moral assessments, frequently resulting in a substantial loss of cultural information. In contrast, language models demonstrate reliable cross-language performance with transfer learning, with LLMs excelling in terms of data efficiency. Importantly, this study also underscores the need for human-in-the-loop validation of automated MF assessment, as even the most advanced models may overlook cultural nuances and face potential risks in cultural misalignment. The findings highlight the potential of LLMs for cross-language MF measurements and other complex multilingual deductive coding tasks.
\end{abstract}

\section{Introduction} 

Moral intuitions have long fascinated social scientists, as they help explain a wide range of cognitive and behavioral phenomena across individuals and groups  \citep{effron2022moral}. Moral foundation theory (MFT) is among the most prominent psychology frameworks for understanding the origin and development of human morality \citep{graham2013moral}. Rooted in moral nativism, MFT argues there are five universal moral foundations---care/harm, fairness/cheating, authority/subversion, loyalty/betrayal, and sanctity/degradation---that transcend languages and cultures and underlie people’s moral judgments and decision-making processes \citep{graham2013moral}.\footnote{Each MF includes both virtue and vice dimensions. We use the virtue label to represent the foundation value.} While some scholars propose other foundations \citep{haidt2012righteous,atari2023morality}, 
these five foundations have received the most empirical validation across domains, languages, and cultures \citep{iurino2020testing}. 

A growing body of literature employs MFT to investigate online social behaviors. For example, MFT provides a framework for understanding rising political polarization. Individuals who prioritize loyalty, authority, and sanctity foundations are more likely to endorse conservative views and engage in polarized political discourse, while those affiliated with liberal ideologies tend to value all MFs more evenly \citep{koleva2012tracing, haidt2007morality}. MFT also sheds light on online cultural clashes. \citet{atran2007religion} identified differing valuations of the sanctity foundation as a key factor in many religious and ideological conflicts. Beyond polarization, MFT has been applied to a range of social issues, including climate change \citep{markowitz2012climate}, vaccine hesitancy \citep{amin2017association}, anti-abortion views \citep{koleva2012tracing}, nationalism \citep{kertzer2014moral}, collective violence \citep{nussio2023moral}, and terrorism \citep{tamborini2020impact}. 



Given its broad relevance, measuring MF values in online discourse is essential; yet, automated extraction of MF values from large-scale texts remains challenging, particularly for non-English corpora. Like other latent human values, MFs are often conveyed through abstract narratives that vary across languages. Also, most computational resources for the measurement of MFs are designed for English content \citep{hoover2020moral, trager2022moral}. This reliance on English hinders cross-cultural comparative research on MFT and limits theoretical advancement from non-English data \citep{cheng2023c}. Although MFT is intended to apply across languages and cultures, the limited availability of non-English resources severely restricts its research scope and further development of the theoretical framework \citep{graham2013moral}.


In this work, we investigate various computational approaches for cross-language measurement of MFs with a particular focus on data-efficiency. We use Chinese as an example and find that (1) MF local language lexicons yield suboptimal performance. They are worse than machine translation approaches that utilize established English measurements such as Mformer. (2) Multilingual encoder-only models can achieve moderate success when trained on annotated English data along with some local language labeled data. However, this strategy is less data-efficient in measuring MFs than in other deductive coding tasks, such as hate speech detection \citep{rottger2022data}. 
(3) Decoder-only LLMs outperform other approaches in cross-language MF measurements in both accuracy and data efficiency. Simply fine-tuning and augmenting with English annotated data can achieve strong performance on non-English corpora; (4) nevertheless, this performance is inconsistent on different MF values, as LLMs may overlook cultural nuances in cross-language measurements, particularly for culturally distinct values \footnote{Code is available on the project's GitHub \url{https://github.com/calvinchengyx/cross-lan-mft-measure}}.

\section{Related Work} 
The unique link between word usage and the expressed moral values provides a theoretical foundation for automated MF measurement from online texts \citep{brady2020mad, gantman2014moral, gantman2016see}. Scholars have explored different computational approaches, including dictionaries \citep{graham2009liberals, hopp2021extended}, word embeddings \citep{kwak2021frameaxis, araque2020moralstrength}, machine learning \citep{lan2022text}, deep learning language models \citep{preniqi2024moralbert, nguyen2024measuring} and LLMs \citep{rathje2024gpt}. These computational methods demonstrate great advantages on scalability and labor intensity compared to traditional human annotations. 

However, these approaches are primarily developed for English, and are not directly applicable to non-English texts due to several major concerns: differences in cultural contexts, the lack of annotated datasets, and limitations in domain generalizability. To address the cross-language challenges and bridge knowledge gaps in MF measurements, various computational approaches have been proposed, which can be broadly categorized into two paths: machine translation to English and the development of cross-language measurement tools \citep{zhuang2020comprehensive}. 

\subsection{English Centric Machine Translation}
Translation is a widely used technique in cross-language MF measurement. For instance, the MFT survey has been translated into over 20 languages for cross-lingual studies \citep{yilmaz2016validation, nilsson2015moral}. For large scale text analysis, the development of multilingual neural machine translation has significantly improved translation quality compared to earlier statistical methods, enhancing context understanding, ambiguity resolution, and fluency \citep{stasimioti2020machine}. This advancement enables a machine-translated approach to cross-language MF measurement by translating target languages into English and applying established English-based methods \citep{artetxe2020translation}.




Established methods for automatically measuring MFs in English include dictionaries \citep{graham2009liberals, frimer2019moral, hopp2021extended}, word embeddings \citep{kwak2021frameaxis, araque2020moralstrength}, machine learning and deep learning models \citep{preniqi2024moralbert, nguyen2024measuring} trained on annotated English-language social media data \citep{hoover2020moral, trager2022moral}. 

\paragraph{Moral foundation dictionaries (MFDs)} 
Word-count methods with crafted English moral lexicons are common. There are four common English MFDs. The original MFD is an expert-crafted dictionary containing a list of 600 words across five foundation values \citep{graham2009liberals}. \citet{frimer2019moral} then expanded this vocabulary to MFD2 to over 2,000 words by automatically identifying similar words with word2vec word embeddings \citep{mikolov2013distributed}. Similarly, \citet{araque2020moralstrength} extended the original MFD to a MoralStrength dictionary with approximately 1,000 English lemmas based on WordNet synsets \citep{wordnet2010princeton}. Compared to MFD2, MoralStrength added a round of crowd-sourced ratings on the expanded lemmas. \citet{hopp2021extended}, however, curated a fully crowd-sourced dictionary named eMFD. It is different from previous expert-curated dictionaries for its layperson focus, contextual annotations, probability labeling and large vocabulary with 3,200 words. 


\paragraph{Moral word embeddings} 
Semantic similarity methods using embeddings are another approach. To address the limitations of word-count methods, such as context insensitivity and vocabulary coverage \citep{nguyen2024measuring}, scholars have introduced semantic similarity methods. For example, \citet{kwak2021frameaxis} proposed an embedding framework FrameAxis. It predefines a vector space of micro moral frames with two sets of opposing seed words. Target documents are then converted into vectors using word embedding models, and their MF values are determined by comparing to the micro-frames. This method has been used to extract MF values from various online texts \citep[e.g.,][]{mokhberian2020moral,jing2021characterizing}. 

\paragraph{Moral language models} 
Supervised classification models with annotated English-language training data have also been used. With advancements in language models and efforts to create human-labeled MF training datasets \citep{hoover2020moral, trager2022moral}, recent studies have demonstrated the potential of fine-tuning language models. For example, \citet{preniqi2024moralbert} fine-tuned a BERT-based classifier MoralBert with large-scale annotated English data and achieved state-of-the-art performance. To address generalizability limitations in out-of-domain datasets \citep{liscio2022cross}, \citet{nguyen2024measuring} proposed another language model Mformer, and reported superior performance compared to other established English methods in evaluations. 


Although machine translation offers several advantages in cross-language MF measurement, including interpretability, scalability, efficiency and accessibility, it also faces significant limitations. First, translation quality varies across languages and domains \citep{ranathunga2023neural}. In some low-resource languages like Tamil, machine translation often makes errors in translating domain terms, polysemous words, and contains repetitions for semantically similar terms \citep{ramesh2021comparing}. Second, it often fails to retain non-propositional information, such as emotional nuances. This can lead to a loss of emotions, toning down, or amplification across languages, introducing bias in subsequent analyses \citep{troiano2020lost}. Third, machine translation struggles to capture cultural elements, which is a major concern for cross-cultural and comparative research \citep{haidt2012righteous}. It often shows limited performance with rare or culture-specific words, idiomatic phrases, and metaphor recognition \citep{dorothy2019lost}. Thus, it remains unclear whether machine translation is a reliable method for measuring cross-language MFs. 

\subsection{Cross-language Measurement Tools}
A second path is to develop cross-language tools, where scholars create computational MF resources tailored to local languages. Common cross-language measurements include local language dictionaries, task-specific encoder-only language models, and LLMs.

\paragraph{Local language dictionaries}
Due to the efficiency at scale and multilingual capabilities, local language dictionaries are widely used to estimate MF values from non-English texts \citep{hopp2021extended}. Developing local language dictionaries generally involves three steps: (1) translating English dictionaries to target languages; (2) adding culturally specific and non-translatable vocabulary; and (3) validating with native speakers and local language corpora.  Several extensive non-English MFDs have been developed and validated in Turkish \citep{alper2020changes}, Japanese \citep{matsuo2019development}, Portuguese \citep{carvalho2020brazilian}, and Chinese \citep{cheng2023c}. Despite the abovementioned advantages, the dictionary approach still faces the inherent limitations of general bag-of-words methods. In this paper, we use C-MFD2---a Chinese MFD---as a cross-language tool to evaluate a local language dictionary approach. We also test a semantic similarity approach using the FrameAxis architecture and cross-language word embedding models. 



\paragraph{Multilingual encoder-only models}
To overcome the limitations of bag-of-words methods, literature has suggested machine learning and deep learning approaches. A primary challenge with these methods is the scarcity of annotated data in local languages for model training \citep[e.g.,][]{ji2024moralbench, nguyen2024measuring}. Therefore, scholars have adopted transfer learning techniques that leverage English-annotated resources for cross-language classifier development. Two major transfer learning strategies are commonly proposed: (1) machine-translating annotated English-language data into local languages to train monolingual models \citep{schuster2019cross}, or (2) using annotated English-language data to train multilingual encoder-only models \citep{barriere2020improving}. Multilingual models have demonstrated strong performance in deductive coding tasks, such as sentiment analysis \citep[e.g.,][]{barriere2020improving} and hate speech detection \citep[e.g.,][]{rottger2022data}. Nevertheless, they also have shown language bias in morality classification tasks, as pre-trained multilingual encoder-only language models often display distinct moral directions across languages \citep{hammerl2022speaking}. This paper focuses on the second transfer learning strategy and evaluates multilingual encoder-only models for cross-language MF measurement.

\paragraph{Large language models (LLMs)}
The rise of decoder-only LLMs provides an alternative for cross-language MF measurement. LLMs show exceptional zero/few-shot learning capability, enabling them to directly label human values out-of-the-box, which is particularly valuable for tasks with limited human annotated data \citep{ziems2024can}. They also have demonstrated strong performance in measuring various human values, including moral reasoning tasks \citep{ziems2024can, agarwal2024ethical}. Notably, LLMs sometimes are not as good as specialized fine-tuned language models \citep{amin2023will,preniqi2024moralbert}, which may be due to the lack of explicit, colloquial definitions of the target human values \citep{ziems2024can}. MFT’s well-established conceptual framework may help address this limitation. Not only are LLMs pre-trained on rich MFT literature \citep[e.g.,][]{abdulhai2023moral}, but MFT also offers clear guidance for crafting clear and effective prompts. Additionally, LLMs trained on vast multilingual data exhibit promising capabilities to handle cross-language measurements \citep{ahuja2023mega}.

Despite the strengths in accessibility, efficiency, multilingualism, and reasoning, there are also some concerns in using LLMs' in cross-language MF measurement. First, LLMs exhibit a substantial degree of subordinate multilingualism, displaying proficiency in some languages but not others \citep{zhang2023don}, which has a strong correlation with the proportion of those languages in the pre-training corpus \citep{li2024quantifying}. Second, there are potential language biases, particularly in human-value relevant coding tasks \citep{kirk2024PRISM, yu2024cmoraleval}. For example, non-English prompts are more likely to generate malicious responses compared to English prompts \citep{shen2024language}. 
Third, different LLMs exhibit varying baseline moral tendencies \citep{ji2024moralbench}. For instance, GPT-3’s MF preferences align more closely with politically conservative individuals when minimal prompt engineering is used in zero-shot learning \citep{abdulhai2023moral}. These moral tendencies, however, are sensitive to prompting, with different prompting strategies significantly influencing classification outcomes in MF measurements \citep{abdulhai2023moral, agarwal2024ethical}.

Thus, it is unclear how LLMs performs in cross-language MF measurement tasks. This paper selects a cutting-edge, open-source model---Llama3.1 \citep{dubey2024llama} to evaluate on LLM approach. 

\section{Data}
Table \ref{tab:datasets} shows details of the benchmarking and training datasets used in this work. We used three human annotated datasets---moral foundation vignettes (MFV), Chinese moral scenarios (CCS) and Chinese core values (CCV), to benchmark the performance of different cross-language MF measurement approaches. We also used three English annotated MFT datasets for model training and fine-tuning.


\begin{table}[ht]
    {\small
    \centering
    \setlength{\tabcolsep}{1.6mm}
    \begin{tabular}{lccccc}
    \toprule
    &\textbf{MFV} &\textbf{CCS} &\textbf{CCV} &\textbf{EN} &\textbf{Total} \\
    \midrule
    care/harm &27 &389 &3,030 &16,607 &20,053 \\
    loyalty/betrayal &16 &248 &1,712 &11,772 &13,748 \\
    authority/subversion &25 &331 &1,278 &13,176 &14,810 \\
    fairness/cheating &12 &259 &1,225 &16,292 &17,788 \\
    sanctity/degradation &10 &226 &247 &9,165 &9,648 \\
    non-moral & 0 & 0 & 0 &28,988 &28,988 \\
    \textbf{Total} &90 &1,453 &7,492 &71,242 &80,277 \\
    \hline
    \end{tabular}
    \caption{Moral foundation annotated datasets used in this paper. MFV, CCS and 20\% of CCV were used for benchmarking; EN and 80\% of CCV were used for fine-tuning XLM-T and Llama3.1-8b language models.}
    \label{tab:datasets}
    }
\end{table}

\paragraph{Moral Foundation Vignettes (MFV)} MFV is a list of social behaviors constructed by psychologists based on MFT, describing the violations of specific moral values from a third-party perspective \citep{clifford2015moral}. It has been widely validated and used to assess measures of moral judgment \citep[e.g.,][]{graham2009liberals,kivikangas2021moral,ji2024moralbench}. We used the MFV as an expert-crafted benchmark stimulus to evaluate the baseline performance of different moral foundation measurement approaches. Since the original MFV is in English, we followed a careful translation process to ensure cultural neutrality in Chinese. First, a native Chinese speaker translated the vignettes with minor modifications to preserve cultural appropriateness. Then, two additional native speakers reviewed the translations to assess whether the scenarios remained representative and meaningful in Chinese cultural contexts. This ensured that the translated vignettes could serve as a culturally neutral benchmark for cross-method comparability.\footnote{The translated vignettes are available on the project's GitHub page.}




\paragraph{Chinese Moral Scenarios (CCS)} CCS is list of moral scenarios written by 202 native Chinese to describe their intuitive understandings of MF values \citep{cheng2023c}. This reverse-annotation method is commonly used for validating MF measurements \citep[e.g.,][]{cheng2023c, frimer2019moral, matsuo2019development}. It incorporates culturally specific content, reflecting native speakers’ natural and intuitive understanding of MFT in a real-word context. 

\paragraph{Chinese Core Values (CCV)} CCV is a human-annotated real-world dataset, including 6,994 sentences collected from four local news websites.\footnote{CCV includes data from the CMOS corpus \citep{peng2021morality}, China Cultural and Ethical Website \url{wenming.cn}, Youth Patriotism News \url{agzy.youth.cn}, and Sohu News \url{news.sohu.com}} The dataset is annotated by three native Chinese speakers based on the Chinese core socialist moral value coding scheme, which is highly correlate with the five universal moral foundation values \citep{liu2022corevalue}. The original CCV dataset includes eight labels,\footnote{The Chinese core socialist moral values include civility, justice, equality, rule of law, patriotism, dedication, integrity, and friendship. Table \ref{fig:cv_mapping} in the Appendix showed the curated mapping scheme.} we employed five Chinese native speakers with postgraduate degrees to re-label the action categories in CCV to five moral foundation values, excluding vice and virtue. The mapped values were determined by majority vote. We sample 20\% of the CCV as the primary benchmarking dataset to evaluate the performance across approaches stratifying on the values. The remainder is used as training data to test the data-efficiency fine-tuning language models.

All benchmarking documents are single-class labeled. For multi-class predicted documents, we decided the measurement performance by a lenient evaluation criterion: a prediction is considered correct if one of the predicted values matches the true label. 


\paragraph{English annotated data (EN)} We use three English annotated MF datasets for transfer learning, including a Twitter corpus \citep{hoover2020moral}, a Reddit corpus \citep{trager2022moral} and a news corpus \citep{hopp2021extended}, which are widely used in training English MF classifiers and show reliable performance \citep{nguyen2024measuring, preniqi2024moralbert}. 

\section{Methods}
\subsection{Machine Translation}
We use Google Translate as an example of a machine translation approach due to its accessibility and consistent performance across domains. 
First, we machine translate benchmarking datasets except MFV to Chinese using the Google Cloud API---Basic Translation service. Then we estimate the MF values from translated documents with established English measurements, including lexicons MFD \citep{graham2009liberals}, MFD2 \citep{frimer2019moral}, eMFD \citep{hopp2021extended} and MoralStrength \citep{araque2020moralstrength}; word embeddings with FrameAxis \citep{kwak2021frameaxis}; and specialized-fine-tuned language models MoralBert \citep{preniqi2024moralbert} and Mformer \citep{nguyen2024measuring}. 

For MFD, MFD2, and eMFD, we calculate word frequencies using the \texttt{eMFDscore} Python package \citep{hopp2021extended}. 
For MFD and MFD 1.0, each document’s MF value is determined by the most frequent MF class in the respective dictionary. A document is mapped to multiple classes if there is an equal number of class matches. If there are no matching words, no class is assigned to the document. eMFD, however, assigns probabilities to its vocabulary, representing their likelihood of being associated with certain MF classes. We sum the probabilities and label the document with the class that has the highest sum. 

We use the \texttt{moralstrength} package \citep{araque2020moralstrength} for the MoralStrength dictionary. We first test its performance of the lexicon features alone with bag-of-word methods; then we train a Support Vector Machine (SVM) model and combine its lexicon features. Since English training sets contain many non-moral labels, while the benchmarking dataset contains only moral labels, we train two SVM models: one with the full training data and another with only moral-labeled training data.

For word embedding methods, we use the \texttt{FrameAxis} Python package \citep{kwak2021frameaxis} to compute anchor micro-frames based on different MFDs with the word2vec embedding model \citep{mikolov2013distributed}. For each MFD, we generate the corresponding micro moral frames from the vocabulary in its class. We then compute and aggregate word contributions to each microframe in the document, and label the document's moral class by identifying significant microframes through comparison with a null model. 

For language models, we apply pre-trained language models \texttt{MoralBert}\footnote{Accessed from \url{https://huggingface.co/vjosap}} and \texttt{Mformers}\footnote{Accessed from \url{https://huggingface.co/joshnguyen}} from HuggingFace with their default settings and no additional fine-tuning. 

\subsection{Local Language Lexicons}
We apply C-MFD2 \citep{cheng2023c} to evaluate the performance of the local language dictionary approach. We test two techniques for locally-developed MF lexicons: word counts and embeddings. For the word embedding method, we test two approaches with the fastText model \citep{grave2018learning}. One involves measure simple semantic similarity. Words in C-MFD2 are grouped into five pseudo-documents based on their MF labels, each serving as anchor frames. MF values are then determined by calculating the semantic distance between the text to be classified and these anchor frames.  
The other uses FrameAxis to construct anchor frames. As C-MFD2 does not have the virtue/vice dimension, which is essential to calculate micro-frames in FrameAxis, we automatically assign this dimension using the RoBERTa-based Chinese sentiment model \texttt{c2-roberta-base-finetuned-dianping-chinese} from HuggingFace.\footnote{Accessed from \url{https://huggingface.co/liam168/c2-roberta-base-finetuned-dianping-chinese}} 

\subsection{Multilingual Encoder-only Models}
We select  XLM-T\footnote{Accessed from \url{https://huggingface.co/cardiffnlp/twitter-xlm-roberta-base}} as the base model to test transfer learning on the multilingual encoder-only model approach \citep{barbieri2022xlm}. Fine-tuned on 198 million multilingual tweets on the XLM-RoBERTa architecture---originally trained on 2.5 TB of Common Crawl data \citep{conneau2019unsupervised}---this model is particularly well-suited for analyzing online content. Also, previous research indicates that it has strong performance in cross-language deductive coding tasks such as hate speech detection, compared to other encoder-only models \citep{rottger2022data}. 

We follow \citet{nguyen2024measuring}'s experience on fine-tuning Mformer with some tweaks. First, we replace the base architecture from RoBERTa-base \citep{liu2019roberta} with \texttt{twitter-xlm-roberta-base} and tokenization is handled by the model's built-in tokenizer \citep{barbieri2022xlm}, with token sequences truncated to a maximum of 512. Second, we set the learning rate, epochs, and batch size to $2e-5$, 3, and 16 respectively following \citet{rottger2022data}. 
Third, we opt to fine-tune five binary classifiers rather than a single multi-label classifier because binary models generally outperform multi-label models in English MF measurements \citep{nguyen2024measuring}.\footnote{We note that in hate speech detection, \citet{rottger2022data} found no significant performance difference between binary and multi-label models when using XLM-T. Given that multi-label models require less storage and training resources, they are a viable alternative for future applications.} Fourth, we adopt a conservative under-sampling strategy in the English annotated training dataset to address class imbalance, establishing a baseline for future improvements. 

After fine-tuning with English annotated data, we further fine-tune each base model with the 80\% of CCV dataset in order to test the data-efficiency as in \citet{rottger2022data}. We incrementally train models with batches of additional CCV data, with each batch containing 100 annotated records.

\subsection{Large Language Models}
For the decoder-only LLMs approach, we select the Llama3.1-8b instruct model, an open-source LLM developed by Meta with a reasonable balance between model performance and computational cost. Compared to closed-source models like GPT-4, Llama3.1 offers greater control, flexibility, transparency, and reproducibility---all of which are important in human-value measurement tasks.

We first test LLMs with prompt-engineering and few-shot learning. Next, we apply the same fine-tuning process used for XLM-T to Llama3.1-8b. Notably, an additional round of data augmentation is performed afterwards, where all English annotations are machine-translated into Chinese using the Google Translate API. Fine-tuning is conducted with the \texttt{unsloth} package using 4-bit quantization on a single NVIDIA L40S GPU. Additionally, we test data efficiency with 20 batches of Chinese annotated items from the CCV dataset. A conservative under-sampling strategy is applied as well with each batch containing 50 records evenly distributed across the five classes.

\subsection{Qualitative Analysis}
We conducted qualitative analysis to gain in-depth understanding of the cultural loss in moral foundation measurements across different approaches. By randomly sampling 100 mislabeled records, we compared predicted labels to ground truth (i.e., human annotated labels), focusing on MFs that are known with cultural distinctions between English and Chinese (i.e., authority, loyalty, and sanctity).

\section{Results}
\subsection{Machine Translation}
As shown in Table \ref{tab:mt}, the performance of machine translation generally fall short in evaluation. With the MFV benchmarking dataset, the lexicon method MFD2 shows the best performance with a weighted F1 score of $0.60$. In the reverse-annotated CCS dataset, a simple SVM model with lexicon features outperforms other measurements ($F1 = 0.74$), but the model coverage is relatively low at only 23\%. In the real-word CCV dataset, the deep learning model Mformer exhibits the strongest performance ($F1 = 0.47$) and maintains comparable results across the other two benchmark datasets. Note that although some MF classes in Mformer displayed good performance (i.e., care/harm, $F1 = 0.72$), some fine-grained MF measurements are very poor. For example, Mformer's prediction on ``loyalty'' ($F1 = 0.21$) is worse than random guessing baseline ($F1 = 0.23$) in the cross-language evaluation setting. 

\begin{table}[!ht]
    {\small
    \centering  
    \setlength{\tabcolsep}{1mm}
    \begin{tabular}{lccccccccc}
        \toprule
        & \textbf{Auth} & \textbf{Care} & \textbf{Fair} & \textbf{Loya} & \textbf{Sanc} & \textbf{Acc} & \textbf{Cov} & \textbf{Fw} & \textbf{Fm} \\
        
        \multicolumn{10}{l}{\textbf{MFV}} \\
        \midrule
        Baseline & 0.28&0.30&0.13&0.18&0.11&0.23&1.00&0.23&0.20 \\
        MFD & \textbf{0.77} & 0.29 & 0.00 & 0.55 & 0.00 & 0.56 & 0.28 & 0.55 & 0.32 \\
        MFD2 & 0.69 & 0.50 & \textbf{0.75} & \textbf{0.59} & 0.33 & \textbf{0.59} & 0.46 & \textbf{0.60} & \textbf{0.57} \\
        eMFD & 0.39 & 0.58 & 0.31 & 0.22 & \textbf{0.44} & 0.42 & \textbf{1.00} & 0.41 & 0.39 \\
        MS & 0.00 & 0.00 & 0.00 & 0.33 & 0.31 & 0.17 & 0.20 & 0.13 & 0.13 \\
        FA+MFD & 0.39 & 0.06 & 0.00 & 0.29 & 0.31 & 0.27 & \textbf{1.00} & 0.21 & 0.21 \\
        FA+MFD2 & 0.19 & 0.18 & 0.06 & 0.43 & 0.17 & 0.22 & \textbf{1.00} & 0.21 & 0.21 \\
        FA+eMFD & 0.07 & 0.47 & 0.19 & 0.40 & 0.17 & 0.31 & \textbf{1.00} & 0.28 & 0.26 \\
        svm+MS & 0.00 & 0.00 & 0.00 & 0.00 & 0.00 & 0.00 & 0.00 & 0.00 & 0.00 \\
        svm+MS* & 0.07 & 0.32 & 0.21 & 0.09 & 0.00 & 0.19 & \textbf{1.00} & 0.16 & 0.14 \\
        MoralBert & 0.27 & 0.55 & 0.50 & 0.10 & 0.30 & 0.38 & \textbf{1.00} & 0.36 & 0.34 \\
        MFormer & 0.65 & \textbf{0.65} & 0.72 & 0.31 & 0.40 & 0.58 & \textbf{1.00} & 0.57 & 0.55 \\
        \midrule
        
        \multicolumn{10}{l}{\textbf{CCS}} \\
        \midrule
        Baseline&0.23&0.27&0.18&0.17&0.16&	0.21&1.00&	0.21&0.22 \\
        MFD &0.47 &0.62 &0.67 &0.18 &\textbf{0.84}&0.54 &0.6 &0.54 &0.55 \\
        MFD2 &0.55 &\textbf{0.71} &0.71 &0.61 &0.71 &0.66 &0.75 & 0.66 &0.66 \\
        eMFD &0.48 &0.52 &0.48 &0.44 &0.25 &0.47 &0.98 &0.45 &0.43 \\
        MS &0.16 &0.11 &0.16 &0.18 &0.21 &0.16 &0.17 & 0.16 &0.16 \\
        FA+MFD &0.45 &0.38 &0.31 &0.26 &0.47 &0.38 &\textbf{1.00} &0.38 &0.37 \\
        FA+MFD2 &0.39 &0.57 &0.57 &0.39 &0.43 &0.47 &\textbf{1.00} &0.47 &0.47 \\
        FA+eMFD &0.33 &0.42 &0.28 &0.17 &0.22 &0.30 &\textbf{1.00} &0.30 &0.28 \\
       svm+MS &\textbf{0.68} &\textbf{0.71} &\textbf{0.77} &\textbf{0.85} &0.61 &\textbf{0.74} &0.23 & \textbf{0.74} &\textbf{0.72} \\
       svm+MS* &0.21 &0.39 &0.28 &0.15 &0.14 &0.26 &\textbf{1.00} &0.25 &0.23 \\
       MoralBert &0.41 &0.63 &0.58 &0.64 &0.49 &0.57 &\textbf{1.00} &0.55 &0.55 \\
       MFormer &0.65 &\textbf{0.71} &0.70 &0.71 &0.71 &0.69 &\textbf{1.00} & 0.69 &0.70 \\
        \midrule
        
        \multicolumn{10}{l}{\textbf{CCV}} \\
        \midrule
        Baseline &0.18&0.40&0.16&0.23&0.03&0.27	&1.00&	0.27&0.20 \\
        MFD &\textbf{0.33} &0.40 &0.43 &0.3 &0.00 &0.34 &0.47 &0.35 &0.29 \\
        MFD2 &0.23 &0.62 &0.43 &0.27 &0.06 &0.43 &0.72 & 0.43 &0.32 \\
        eMFD &0.11 &0.59 &0.36 &0.20 &0.02 &0.40 &\textbf{1.00} &0.36 &0.26 \\
        MS &0.17 &0.21 &0.23 &0.23 &0.09 &0.20 &0.36 & 0.21 &0.19 \\
        FA+MFD &0.29 &0.30 &0.27 &\textbf{0.31} &0.06 &0.28 &\textbf{1.00} &0.29 &0.25 \\
        FA+MFD2 &0.07 &0.49 &0.35 &0.30 &0.11 &0.34 &\textbf{1.00} &0.34 &0.27 \\
        FA+eMFD &0.04 &0.56 &0.31 &0.24 &0.08 &0.38 &\textbf{1.00} &0.34 &0.25 \\
       svm+MS &0.28 &0.59 &0.49 &0.27 &\textbf{0.18 }&0.47 &0.17 & 0.44 &\textbf{0.36 }\\
       svm+MS* &0.12 &0.35 &0.25 &0.12 &0.03 &0.23 &\textbf{1.00} &0.23 &0.17 \\
       MoralBert &0.15 &0.62 &0.47 &0.24 &0.10 &0.45 &\textbf{1.00} &0.41 &0.32 \\
       MFormer &0.23 &\textbf{0.72} &\textbf{0.52} &0.21 &0.14 &\textbf{0.50} &\textbf{1.00} & \textbf{0.47} &\textbf{0.36} \\
        \bottomrule
    \end{tabular}
    }
    \caption{Established English moral foundation measurements applied to machine translated text. Acc, Cov, Fw and Fm refers to accuracy, coverage, F1 weighted and F1 macro respectively. Baseline, FA and MS in the first column represent random guessing, FrameAxis and MoralStrength. The best performing methods for each dataset are in \textbf{bold}. MoralStrength models trained with no non-moral labeled data are marked with a star (*). The Coverage column measures the percentage of all moral labels in the prediction.}
    \label{tab:mt}
\end{table}

\subsection{Local Language Lexicons}
Local language lexicon approaches demonstrate similarly moderate performance with word-count methods. Table \ref{tab:lld-dict} shows that across all benchmarking datasets, C-MFD2 consistently outperforms other methods although its performance is generally still poor. In the real-world CCV dataset, C-MFD2 achieves the best performance ($F1 = 0.43$), which is comparable to the machine translation approach with Mformer. We find multilingual word embedding model fastText with FrameAxis framework do not improve the cross-language measurement performance in this task. 



\begin{table}[!ht]
    {\small
    \centering  
    \setlength{\tabcolsep}{1mm}
    \begin{tabular}{lccccccccc}
        \toprule
        & \textbf{Auth} & \textbf{Care} & \textbf{Fair} & \textbf{Loya} & \textbf{Sanc} & \textbf{Acc} & \textbf{Cov} & \textbf{Fw} & \textbf{Fm} \\
        
        \multicolumn{10}{l}{\textbf{MFV}} \\
        \midrule
        Baseline & 0.28&\textbf{0.30}&0.13&0.18&	0.11&0.23&1.00&0.23&0.20 \\
        cMFD2 &\textbf{0.48 }&0.18 &\textbf{0.67} &0.29 &\textbf{0.22} &\textbf{0.39} &0.40 & \textbf{0.42} &\textbf{0.37} \\
        FT &0.44 &0.00 &0.00 &0.12 &0.18 &0.30 &\textbf{1.00} &0.16 &0.15 \\
        FT+FA &0.08 &0.00 &0.00 &\textbf{0.30 }&0.00 &0.19 &\textbf{1.00} & 0.08 &0.08 \\

        \midrule
        \multicolumn{10}{l}{\textbf{CCS}} \\
        \midrule
        Baseline&0.23&0.27&0.18&0.17&0.16&	0.21&1.00&	0.21&0.22 \\
        cMFD2 &\textbf{0.74} &\textbf{0.76} &\textbf{0.73} &\textbf{0.68} &\textbf{0.74} &\textbf{0.74} & 0.76 & \textbf{0.73} &\textbf{0.73} \\
        FT&0.44 &0.32 &0.35 &0.39 &0.37 &0.39 &\textbf{0.97} &0.37 &0.37 \\
        FT+FA &0.14 &0.05 &0.15 &0.29 &0.02 &0.20 &\textbf{0.97} &0.12 &0.13 \\
        
        \midrule
        \multicolumn{10}{l}{\textbf{CCV}} \\
        \midrule
        Baseline &0.18&0.40&0.16&0.23&0.03&0.27	&1.00&	0.27&0.20 \\
        cMFD2 &0.26 &\textbf{0.63} &\textbf{0.45} &0.28 &\textbf{0.09} &\textbf{0.45} &0.64 & \textbf{0.43} &\textbf{0.34} \\
        FT &\textbf{0.28} &0.17 &0.09 &0.01 &0.03 &0.20 &\textbf{0.98} &0.13 &0.12 \\
        FT+FA &0.02 &0.03 &0.04 &\textbf{0.37} &0.00 &0.23 &\textbf{0.98} & 0.10&0.09 \\
    \bottomrule
    \end{tabular}
    }
    \caption{The performance of C-MFD2 in cross-language MF measurement. FT and FA in the first column represent FastText and FrameAxis. The best performing methods for each dataset are in \textbf{bold}. Other abbreviations are the same as those in Table \ref{tab:mt}.}
    \label{tab:lld-dict}
\end{table}

\subsection{Multilingual Encoder-only Models}
The multilingual encoder-only model XLM-T, trained on the same English annotated data, shows a moderate but reduced performance compared to the monolingual English model Mformer. In Table \ref{tab:xlm}, XLM-T shows moderate performance across all benchmarking datasets ($F1_{MFV} = 0.71, F1_{CCS} = 0.71, F1_{CCV} = 0.63$) outperforming both machine translation and local lexicon approaches. In addition, its performance is generally consistent across the five foundation values, showing a better reliability in the fine-grained MF measurements. This consistency is likely due to training five separate classifiers. In the real-world CCV dataset, the fine-tuned XLM-T model has moderate performance in four out of five foundation values with an average F1 score of $0.63$. The ``authority'' model has lower performance with $F1 = 0.46$.

\begin{table}[!ht]
    {\small
    \centering  
    \begin{tabular}{lcccccc} 
    \toprule
    &\textbf{Auth} &\textbf{Care} &\textbf{Fair}  &\textbf{Loya} &\textbf{Sanc} & \textbf{XLM-T Avg} \\
    \multicolumn{6}{l}{\textbf{MFV}} \\
    \midrule
    0 &0.70 &0.81 &0.76 &0.88 &0.85 & 0.80\\
    1 &0.27 &0.41 &0.15 &0.54 &0.15 & 0.30\\
    Acc &0.58 &0.71 &0.62 &0.81 &0.74 & 0.69\\
    Fm &0.49 &0.61 &0.45 &0.71 &0.50 & 0.55\\
    Fw &0.58 &0.69 &0.68 &0.82 &0.77 & 0.71 \\

    \midrule
    \multicolumn{6}{l}{\textbf{CCS}}\\
    \midrule
    0 &0.66 &0.76 &0.79 &0.91 &0.89 & 0.80\\
    1 &0.19 &0.44 &0.02 &0.54 &0.51 & 0.34\\
    Acc &0.52 &0.66 &0.65 &0.85 &0.82 & 0.70 \\
    Fm &0.42 &0.60 &0.4 &0.73 &0.70 & 0.57\\
    Fw &0.55 &0.67 &0.65 &0.85 &0.83 & 0.71\\
    
    \midrule
    \multicolumn{6}{l}{\textbf{CCV}} \\ 
    \midrule
    0 &0.51 &0.63 &0.74 &0.78 &0.75 & 0.68\\
    1 &0.24 &0.58 &0.20 &0.41 &0.10 & 0.31\\
    Acc &0.40 &0.60 &0.61 &0.68 &0.61 &0.58 \\
    Fm &0.37 &0.60 &0.47 &0.60 &0.43 &0.49\\
    Fw &0.46 &0.61 &0.65 &0.70 &0.73 & 0.63\\
    \bottomrule
    \end{tabular}
    }
    \caption{Multilingual encoder-only models for moral foundation measurement. \textit{XLM-T Avg} refers to the average F1 score across five foundation models.}
    \label{tab:xlm}
\end{table}

\begin{figure}[t]
\centering
\includegraphics[width=\columnwidth]{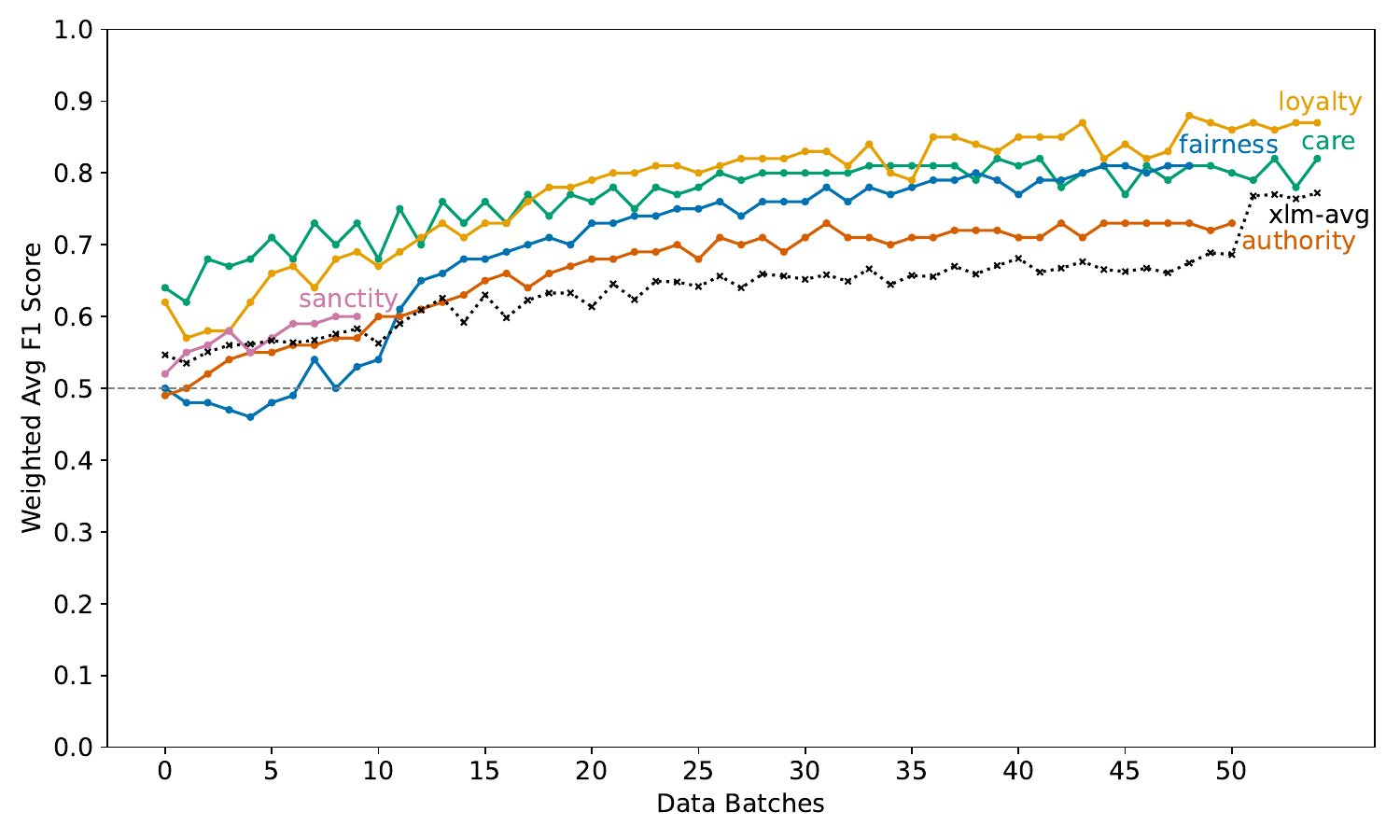} 
\caption{Accumulated fine-tuning of the XLM-T with local language annotated data from the CCV dataset. Each batch includes 100 items.}
\label{fig:xlm-batch-cv}
\end{figure}

In the follow-up evaluation of batch training with local language data, we observe improved model performance as the volume of local language training data increases. As shown in Figure \ref{fig:xlm-batch-cv}, the fine-tuned XLM-T models achieve higher F1 scores with more locally labeled data. For example, the F1 score of the ``loyalty'' model rose from $0.62$ to $0.80$ with 22 batches of local-language data. Similar increasing patterns are observed across all five models. 

However, the amount of annotated non-English data required to reach reliable classification thresholds for MFs is much more than for hate speech detection \citep{rottger2022data}. On average, 16 batches are needed to reach the F1 score of $0.70$ and 49 batches to reach $0.80$. The ``care'' model, which is the best performing model of the five, requires six additional batches to reach an F1 score of $0.70$ and 27 batches to reach $0.80$. There are 10 batches of data available for the ``sanctity'' model, and it shows no significant improvement even once all 10 batches are used for training.

It is worth noting that limited local language annotated data may also negatively impact the model’s performance. For example, the ``fairness'' model's F1 score initially drops from $0.50$ to $0.46$ when fine-tuned with local-language data. It only begins to stabilize and improve after 11 batches. In summary, while moderate performance can be achieved with local-language annotations, considerably more data is required to attain robust classification performance with the multilingual encoder-only model approach in cross-language MF measurements.

\begin{table}[tb]
    {\small 
    \centering  
    \setlength{\tabcolsep}{1mm}
    \begin{tabular}{lcccccccccc}
    \toprule
    &\textbf{Auth} &\textbf{Care} &\textbf{Fair} &\textbf{Loya} &\textbf{Sanc} &\textbf{Acc} &\textbf{Cov} &\textbf{Fw} &\textbf{Fm} \\
    \textbf{MFV} & & & & & & & & & \\
    \midrule
    Baseline &0.28 &0.30 &0.13 &0.18 &0.11 &0.23 &1.00 &0.23 &0.20 \\
    en$\times\varnothing$ &0.41 &0.78 &0.40 &0.35 &\textbf{0.40 }&0.55 &\textbf{0.61 }&0.54 &0.47 \\
    zh$\times\varnothing$ &0.56 &0.40 &0.25 &0.00&0.00&0.42 &0.36 &0.34 &0.24 \\
    en$\times$en &0.89 &0.87 &0.94 &0.57 &0.00&0.84 &0.41 &0.81 &0.65 \\
    zh$\times$en &0.82 &0.88 &0.75 &0.67 &0.00&0.79 &0.32 &0.77 &0.62 \\
    en$\times$zh &0.89 &0.86 &0.90 &\textbf{0.80} &0.00 &0.86 &0.54 &0.85 &0.69 \\
    zh$\times$zh &0.91 &0.83 &0.91 &0.75 &0.50 &0.85 &0.58 &0.84 &\textbf{0.78} \\
    en$\times$(en+zh) &0.88 &\textbf{0.89 }&\textbf{0.95} &\textbf{0.80} &0.00&\textbf{0.87} &0.60 & \textbf{0.86} &0.70 \\
    zh$\times$(en+zh) &\textbf{0.93} &0.88 &0.91 &0.77 &0.00&\textbf{0.87 }&0.60 &0.85 &0.70 \\
    \midrule
    \textbf{CCS} & & & & & & & & & \\
    \midrule
    Baseline &0.23 &0.27 &0.18 &0.17 &0.16 &0.21 &1.00 &0.21 &0.22 \\
    en$\times\varnothing$ &0.58 &0.78 &0.71 &0.64 &0.71 &0.70 &\textbf{0.93} & 0.69 &0.68 \\
    zh$\times\varnothing$ &0.66 &0.80 &0.71 &0.63 &\textbf{0.72} &0.71 &0.82 &0.71 &0.71 \\
    en$\times$en &\textbf{0.78} &\textbf{0.85} &\textbf{0.88} &0.84 &0.68 &\textbf{0.82 }&0.52 & \textbf{0.82} &0.80 \\
    zh$\times$en &0.75 &0.84 &\textbf{0.88} &0.80 &0.69 &0.80 &0.45 &0.80 &0.79 \\
    en$\times$zh &0.73 &\textbf{0.85} &0.84 &0.85 &\textbf{0.72} &0.81 &0.56 &0.80 &0.80 \\
    zh$\times$zh &0.74 &0.82 &0.85 &0.86 &0.65 &0.80 &0.56 &0.80 &0.78 \\
    en$\times$(en+zh) &0.76 &\textbf{0.85} &0.86 &\textbf{0.87} &0.69 &\textbf{0.82} &0.55 & \textbf{0.82} &\textbf{0.81} \\
    zh$\times$(en+zh) &0.74 &0.84 &0.85 &\textbf{0.87} &0.70 &0.81 &0.57 &0.81 &0.80 \\
    \midrule
    \textbf{CCV} & & & & & & & & & \\
    \midrule
    Baseline &0.18 &0.40 &0.16 &0.23 &0.03 &0.27 &1.00 &0.27 &0.20 \\
    en$\times\varnothing$ &0.29 &0.69 &0.47 &0.50 &0.06 &0.54 &\textbf{0.80} &0.53 &0.40 \\
    zh$\times\varnothing$ &0.29 &0.70 &0.35 &0.16 &0.13 &0.48 &0.61 &0.42 &0.32 \\
    en$\times$en &0.28 &0.83 &0.68 &0.55 &0.00 &0.66 &0.62 &0.65 &0.47 \\
    zh$\times$en &0.19 &0.82 &0.59 &0.23 &0.00 &0.59 &0.41 &0.57 &0.37 \\
    en$\times$zh &0.39 &\textbf{0.85} &0.77 &0.69 &0.36 &0.73 &0.68 &0.72 &0.61 \\
    zh$\times$zh &0.32 &0.81 &0.73 &0.70 &0.36 &0.71 &0.69 &0.69 &0.59 \\
    en$\times$(en+zh) &\textbf{0.42} &\textbf{0.85} &\textbf{0.79} &\textbf{0.72 }&\textbf{0.43 }&\textbf{0.75} &0.72 &\textbf{0.74} &\textbf{0.64} \\
    zh$\times$(en+zh) &0.36 &0.84 &0.75 &\textbf{0.72} &0.31 &0.73 &0.72 &0.72 &0.60 \\
    \bottomrule
    \end{tabular}
    }
    \caption{Llama3.1-8b model with few-shot learning prompts for cross-language moral foundation measurements. The first language refers to the language of prompt while the languages after $\times$ refer to the language(s) of fine-tuning datasets. $\varnothing$ denotes the empty set when no fine-tuning is done. For example, zh$\times$(en+zh) denotes a Chinese prompt on a model fine-tuned using English and Chinese data. The best performing method per dataset is in \textbf{bold}. Column labels are consistent with Table \ref{tab:mt}.}
    \label{tab:llama-8b}
\end{table}

\begin{table}[!ht]
    {\small 
    \centering  
    \setlength{\tabcolsep}{1mm}
    \begin{tabular}{lcccccccccc}
    \toprule
    &\textbf{Auth} &\textbf{Care} &\textbf{Fair} &\textbf{Loya} &\textbf{Sanc} &\textbf{Acc} &\textbf{Cov} &\textbf{Fw} &\textbf{Fm} \\
    \textbf{MFV} & & & & & & & & & \\\midrule
    Baseline &0.28 &0.30 &0.13 &0.18 &0.11 &0.23 &1.00 &0.23 &0.20 \\
    8b en  &0.41 &0.78 &0.40 &0.35 &0.40 &0.55 &0.61&0.54 &0.47 \\
    8b zh &0.56 &0.40 &0.25 &0.00 &0.00 &0.42 &0.36 & 0.34 &0.24 \\
    70b en  &0.67 &\textbf{0.82} &0.52 &0.48 &0.55 &\textbf{0.66} &\textbf{0.83 }&\textbf{0.66} &0.61 \\
    70b zh &\textbf{0.73} &0.67 &\textbf{0.53} &\textbf{0.56} &\textbf{0.60} &0.63 &0.64 &0.64 &\textbf{0.62} \\
    \midrule
    \textbf{CCS} & & & & & & & & & \\\midrule
    Baseline &0.23 &0.27 &0.18 &0.17 &0.16 &0.21 &1.00 &0.21 &0.22 \\
    8b en  &0.58 &0.78 &\textbf{0.71} &0.64 &0.71 &0.70 &0.93 &0.69 &0.68 \\
    8b zh  &\textbf{0.66} &0.80 &\textbf{0.71} &0.63 &\textbf{0.72} &\textbf{0.71} &0.82 &\textbf{0.71} &\textbf{0.71} \\
    70b en &0.54 &0.77 &0.63 &0.68 &0.65 &0.67 &\textbf{0.96} &0.66 &0.65 \\
    70b zh  &0.56 &\textbf{0.83 }&0.63 &\textbf{0.71 }&0.70 &0.70 &0.89 &0.69 &0.69 \\
    \midrule
    \textbf{CCV} & & & & & & & & & \\
    \midrule
    Baseline &0.18 &0.40 &0.16 &0.23 &0.03 &0.27 &1.00 &0.27 &0.20 \\
    8b en  &\textbf{0.29} &0.69 &0.47 &0.50 &0.06 &0.54 &0.80 &0.53 &0.40 \\
    8b zh  &\textbf{0.29} &0.70 &0.35 &0.16 &0.13 &0.48 &0.61 &0.42 &0.32 \\
    70b en  &0.18 &0.69 &0.52 &0.41 &0.15 &0.55 &\textbf{0.86} &0.51 &0.39 \\
    70b zh &0.18 &\textbf{0.76} &\textbf{0.54} &\textbf{0.59} &\textbf{0.18} &\textbf{0.62} &0.63 &\textbf{0.60} &\textbf{0.45} \\
    \bottomrule
    \end{tabular}
    
    }
    \caption{\textit{Llama3.1} 8b and 70b models with few-shot prompting for cross-language moral foundation measurement. The best performing method for each dataset is in \textbf{bold}. Column labels are consistent with Table \ref{tab:mt}.}
    \label{tab:llama-70b}
\end{table}

\subsection{Large Language Models}
Table \ref{tab:llama-8b} shows the results of the LLMs approach. With only prompt engineering, Llama3.1-8b already performs better than the machine translation and local language lexicon approaches across all benchmarking datasets. After fine-tuning with annotated English-language data, it immediately achieves strong performance on the MFV ($F1 = 0.81$) and CCS ($F1 = 0.82$) datasets, and moderate performance on the CCV ($F1 = 0.65$) dataset. This performance exceeds that of XLM-T with the same English-language training data ($F1 = 0.63$). 

The model is fine-tuned with a data augmentation strategy where the English labeled data are machine translated into Chinese and fed into the LLM again. This step significantly improves the model performance and results in the best performance for all three benchmarking dataset (MFV $F1 = 0.86$; CCS $F1 = 0.82$; CCV $F1 = 0.74$). These F1 scores exceed all other approaches tested in this paper. 

Additionally, we notice that English prompts generally outperform non-English prompts on Llama3.1-8b, even when analyzing non-English documents. This difference is more pronounced in the base instruct model but becomes less significant with progressive fine-tuning using annotated data. After fine-tuning with English data, the difference in the F1 scores for English and Chinese prompts on the CCV dataset decreases from 0.11 to 0.08, further reducing to 0.03 with data augmentation and 0.02 when fine-tuned with both English and translated Chinese data.


Moreover, a better LLM base model may further improve the model performance. Table \ref{tab:llama-70b} shows that under the same condition, Llama3.1-70b outperforms Llama3.1-8b when prompting in local languages. In the CCV dataset, under a few-shot learning setup with Chinese prompting, the larger LLM ($F1 = 0.60$) is better than both its smaller counterpart ($F1 = 0.42$) and the English prompting counterpart ($F1 = 0.51$).

%

%
%

%
%

\subsection{Qualitative Analysis of Cultural Nuances}
We selected MFormer and fine-tuned Llama3.1-8b-instruct with English and translated Chinese data as models to analyze cultural loss in translation and non-translation approaches. We identified three common types of information loss during the translation process (See Table \ref{fig:quali} in the Appendix). First, idioms and slang are often mistranslated, as in Example 1, where the Chinese phrase meaning ``faked or staged accident for compensation'' is simply translated to its superficial meaning. Second, contextual meaning. As illustrated in Example 2, where the English translation ``look towards'' misses the cultural connotation of ``favoring someone,'' resulting in different expressed moral foundations. Third, political euphemisms lose their cultural context, as in Example 3, where the English translation fails to convey that the text refers to civil servant behavior. These information loss in the translation process often result in mislabeling in MF measurement. 

We further analyzed the non-translational approach to understand how LLMs process cultural nuances. Compared to the translational approach, LLMs better recognize slang and idioms (Examples 2 and 13), but demonstrate more subtle cultural misalignment. Our analysis revealed several distinct patterns. Most notably, LLMs failed to distinguish certain Confucian morality concepts such as filial piety, which encompass family responsibilities to respect and care for elders (Examples 6, 8, 9, and 10). Native Chinese speakers tend to labeled these instances as ``authority,'' whereas language models often only categorized them as ``care.'' Additionally, we observed misalignment regarding government and local political entities. Human annotators tended to label cases involving government (Examples 3, 4) and officials (Example 11) as ``loyalty,'' while LLMs classified them as ``authority'' or ``fairness.'' Finally, examining misalignment in ``sanctity'' revealed that Chinese annotators frequently associated sanctity with corrupt politicians or role model civil servants (Examples 11 and 12), a distinction rarely captured by LLMs.

\color{black}
\section{Discussion and Limitations} 



We find traditional approaches, such as machine translation and local language lexicons, may not be the most effective solutions for cross-language moral foundation measurements. Instead, fine-tuning multilingual encoder-only models and leveraging LLMs through transfer learning emerge as promising alternatives. Notably, LLMs demonstrate strong performance and data efficiency, making them particularly well-suited for addressing the challenges of cross-language MF analysis. 

Above all, simple machine translation approach has proven suboptimal in cross-language MF measurement. This issue is particularly concerning for measuring culturally significant values. Machine translation often introduces bias in the rendering of slangs, contextual meanings, and political euphemisms into local languages. Moreover, when relying on pretrained English classifiers such as MFormer—which is predominantly trained on English-annotated data—this cultural information loss may be further amplified. For instance, MFormer’s classification performance on the “loyalty” foundation in translated data ($F1 = 0.21$) falls below random chance ($F1 = 0.23$), indicating a substantial loss of cultural nuance. Nevertheless, it is important to note that machine translation models are rapidly evolving. Current deficiencies—such as loss of contextual nuance or culturally embedded moral cues—limit their effectiveness in this domain. Advances in context-aware and domain-adapted translation models \citep{jin2023challenges,saunders2022domain}, particularly those fine-tuned on moral discourse, may mitigate some of these issues, which deserves future exploration.

Researchers should also be cautious with local language lexicons---they perform worse than the machine translation for cross-language MF measurement in some cases. Given the inherent limitations of lexicon-based methods and the extensive resources required to develop them, they are inefficient for cross-language MF measurement, particularly at the document level where semantic complexity is higher. However, culturally specific moral lexicons may still provide valuable insights at the vocabulary level for cross-cultural research, though broader applications should be approached with careful validations.

Fine-tuning LLMs is more data-efficient than fine-tuning multilingual encoder-only models for cross-language MF measurement. In the CCV benchmarking dataset, XLM-T requires on average more than 2,000 local-language annotated records (20 batches) to achieve strong performance ($F1 = 0.75$). In contrast, Llama3.1-8b can reach comparable performance with only English data machine-translated to Chinese and thus no local language labeling. Our findings also suggest that larger LLMs such as Llama3.1-70b may perform better for measuring MFs in non-English corpora. That is, updating the base LLM to a more power model with enhanced multilingual capabilities could further improve the performance of the LLM approach.

Nevertheless, several limitations of using LLMs for cross-language MF measurement should be acknowledged. Firstly, the performance on fine-grained MF values can vary. Although the measurement on some culturally sensitive values like ``loyalty'' is relatively reliable, others like ``authority'' and ``sanctity'' still miss significant cultural nuances. 

\begin{figure}[t]
\centering
\includegraphics[width=\columnwidth]{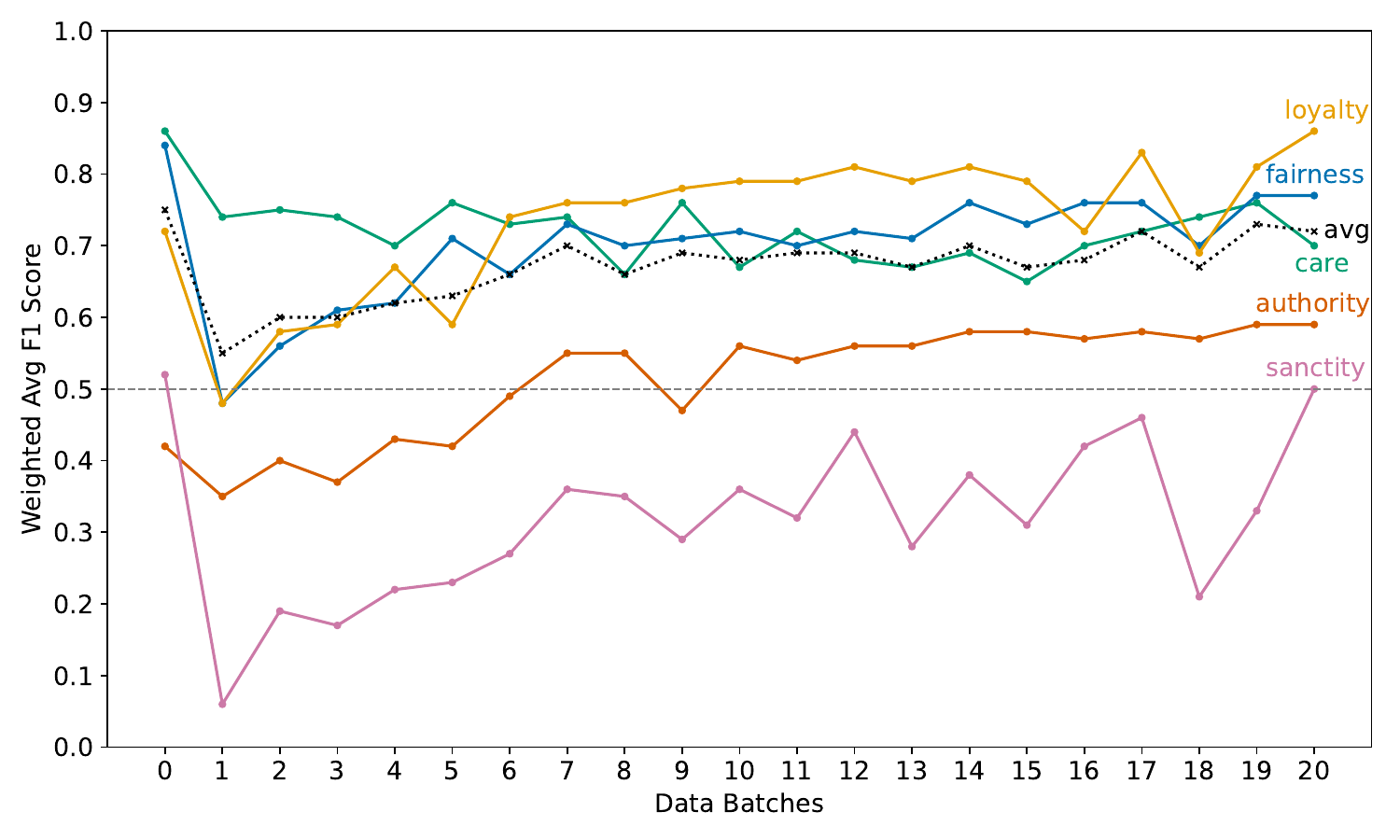} 
\caption{Accumulated fine-tuning Llama3.1-8b with local language annotated data from the Chinese CV dataset. Each batch includes 50 items evenly distributed across five classes.}
\label{fig:llama-batch-cv}
\end{figure}

Second, fine-tuning LLMs might backfire. Fine-tuning with limited local-language data may degrades model performance. While training with local language annotated data is commonly used to incorporate cultural specificity and enhance transfer learning, this approach appears less effective for LLMs in MF measurements task when data volume is small. We fine-tuned Llama3.1-8b using 20 batches of CCV data with the same strategy used in XLM-T. Each batch containing 50 MF labels evenly distributed across five classes. As shown in Figure \ref{fig:llama-batch-cv}, the initial drop in model performance struggles to recover within the available data for "care", "fairness", and "sanctity" values. 

In addition, fine-tuning with English-annotated data may introduce extra cultural bias to LLMs. We replicated the same finetuning strategy to an Italian dataset ``moralConvITA'' \citep{stranisci2021expression}, and observed similar significant performance drops in culturally-nuanced values like "loyalty" and "sanctity" (see Appendix). This signals the risk of cultural bias in English-centric LLMs. Finetuning such LLMs with English annotation data may exacerbate their bias against non-English languages, resulting in more cultural information loss in the measurement task, which echoes concerns about LLMs being shaped by values associated with WEIRD populations \citep{atari2023humans}. These findings highlight the need for culturally-sensitive fine-tuning approaches that preserve cross-cultural moral nuances.



Third, we note that our experiments were conducted in a single non-English language, which may limit the generalizability of our findings. Nevertheless, we hope this work can serve as a foundation for future research that extends to a broader range of non-English languages on MF measurement, particularly low-resource ones, where such methodological advancements would be especially valuable.



Finally, we highlight the essential role of human validation in measuring cross-language MFs with LLMs. Given the limitations discussed, human validation is strongly recommended to ensure the reliability of LLM-based cross-language MF measurements. LLMs can effectively assist in this process by providing rationales alongside their classification outputs, supporting human evaluators and facilitating a more efficient result assessment.



\section{Conclusions}

This study examined four computational approaches---machine translation, dictionary, multilingual encoder-only models and decoder-only LLMs---for automatically measuring moral foundation values in non-English texts. It uses Chinese as a case study and leverages established English resources for this cross-language deductive coding task. 

It first highlights the limitations of dictionary and machine translation approaches: while local language dictionaries can support lexicon-level analysis, they often lack the depth needed for complex semantic assessments. Notably, advanced English-based tools applied to machine-translated data outperform local lexicon-based methods, which underscores the limitations of lexicon approaches in this task. The study then explores the potential of transfer learning in language models, showing that both multilingual encoder-only models and LLMs demonstrate strong performance, with LLMs performing better and being more data-efficient. It is recommended to select approaches based on the availability of local-language annotated data. When sufficient data is available, a smaller multilingual language model generally yields satisfactory results. Otherwise, LLMs can serve as a reliable tool.

We recommend the following steps for applying LLMs to cross-language MF measurement: (1) start with multilingual LLMs and use culturally specific prompt engineering; (2) adopt a binary classification approach for each moral foundation; (3) fine-tune models using available English annotations alongside carefully translated local-language data, while curating out English-centric value cases; and (4) incorporate human validation, especially for culturally distinctive values.

Findings in this paper provide valuable insights for cross-cultural MF research and shed light on future applications of LLM-assisted deductive coding in multilingual tasks.

{\small
\bibliography{reference}
}

\section{Appendix} 

\subsection{Methods}
We provide additional details on model training and fine-tuning beyond those reported in the method section. Python scripts for replication are available on GitHub. All models were trained and fine-tuned on identical English annotation datasets: moral foundation Reddit corpus, moral foundation Twitter corpus, and extended moral foundation dictionary news corpus.\footnote{MFRC \url{https://paperswithcode.com/dataset/mfrc}, MFTC \url{https://osf.io/k5n7y/}, eMFD \url{https://osf.io/preprints/psyarxiv/924gq_v1}} Computations were performed on a single NVIDIA L40S GPU.

\subsubsection{XLM-T}
Task objective: Binary classification of five moral foundation values. Base model: \texttt{cardiffnlp/twitter-xlm-roberta-base} from Hugging Face. Hyperparameters: Learning rate: $2e-5$. Epochs: 3. Batch size: 16 (train \& eval). Weight decay: 0.01. Warmup steps: 100.

\subsubsection{Llama3.1}
Task objective: Multiclass classification of five moral foundation values. Base model: \texttt{llama3.1-8b-instruct} from Hugging Face. Framework: Unsloth with PEFT via LoRA. Precision: LoRA applied with 4-bit quantization. We used the llama-3 chat template from Unsloth. Hyperparameters: Learning rate: $2e-5$. Epochs: 3. Batch size: 128 (train \& eval). Weight decay: 0.01. Max Seq Length: 1024. LoRA configurations used default settings: r-16, targeting q, k, v, o, gate, up, and down-proj parameters. Random state was set to 3047. The English system prompt is in the Appendix. Chinese and Italian prompts were translated accordingly and shared on the GitHub. Few-shot examples ($N=3$) were purposefully sampled from the benchmarking dataset in the local language.

\subsection{Results}

\subsubsection{Qualitative Analysis}
\begin{figure}[!ht]
    \centering
    \includegraphics[width=0.8\pdfpagewidth]{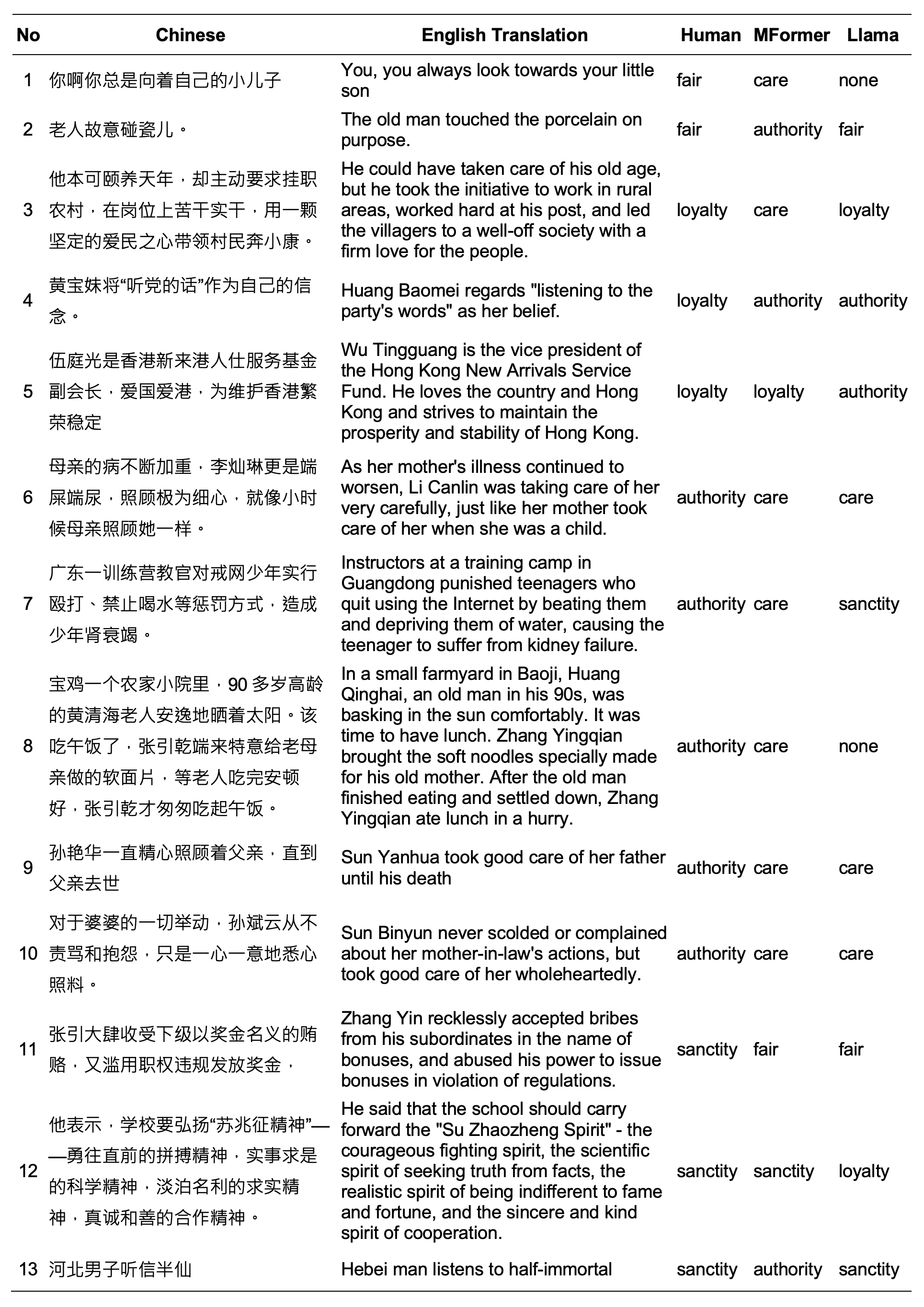}
    \caption{Qualitative analysis of cultural nuances. \textit{Chinese} column refers to the target Chinese text; \textit{English Translation} column is the result of machine translation using Google Translate. \textit{Human, MFormer and Llama} refer to the labels annotated by human annotators, MFormer, and Llama3.1-8b-instruct model. }
    \label{fig:quali}
\end{figure}

Table \ref{fig:quali} shows examples of mis-annotated texts in the CCV dataset. English translations were generated via Google Translate using the \texttt{deep-translator} Python package. MFormer shows results from machine-translated English text, while Llama shows results from original Chinese texts. We used the Llama3.1-8b-instruct model fine-tuned on English annotations and their Chinese translations.

\subsubsection{LLMs Approach on Italian}

We tested a similar LLM-based strategy on an Italian dataset ``moralConvITA'' \citep{stranisci2021expression}, with results shown in Table \ref{tab:italian}. This benchmark comprises 1,724 unique Twitter posts discussing the immigration issue in Italy, labeled by native Italian speakers ($N=8$) according to five moral foundations. To minimize annotation ambiguity, we sampled approximately 100 texts with distinct labels for each moral foundation.

We applied the same supervised instructive fine-tuning strategy used in the CCV dataset. First, we used base Llama3.1-8b-instruct model with both English and Italian prompts. Then, we fine-tuned the model with English annotated text-label pairs from Twitter, Reddit, and news corpora used on CCV. Finally, we performed data augmentation using Italian translations generated via Google Translate. We maintained similar default settings with learning rate, epochs, and batch size at 2$e$-5, 1, and 16, respectively.

We observed similar results as shown in CCV with some noticeable distinctions. Finetuning with translated local language annotations shows strong potential in enhancing LLMs’ performance on culturally distinct MF values in CCV. For example, Llama3.1-8b notably improves on ``loyalty'' when fine-tuned with English data machine-translated to Chinese ($F1 = 0.69$), compared to its performance with the original English training data ($F1 = 0.57$). Such strategy may serve as a practical and cost-effective way to improve model performance in cross-language measurements, especially in cases where mass-labeling of local-language data is infeasible due to the resource-intensive process of creating high-quality language-specific labeled datasets.

This strategy, however, showed moderate performance on the Italian dataset. This difference may be attributed to the smaller Italian benchmark sample size ($N = 492$ vs. approximately 1,500 in CCV), the unstructured nature of Twitter data (typically brief with limited context), and differences in domain and language complexity. The brevity and informality of Italian tweets likely posed greater challenges for accurate moral foundation inference.

\begin{table}[!ht]
    {\small 
    \centering  
    \setlength{\tabcolsep}{1mm}
    \begin{tabular}{lcccccccccc}
    \toprule
    &Auth &Care &Fair &Loya &Sanc &Acc &Cov &Fw &Fm \\\midrule
    baseline &0.20 &0.20 &0.20 &0.20 &0.20 &0.20 &1.00 &0.20 &0.20 \\
    en$\times\varnothing$ &0.53 &\textbf{0.52} &\textbf{0.43} &\textbf{0.57} &0.26 &\textbf{0.48} &0.75 &\textbf{0.46} &\textbf{0.46} \\
    it$\times\varnothing$ &0.54 &0.44 &0.27 &0.35 &0.30 &0.38 &0.39 &0.38 &0.38 \\
    en $\times$ en &0.57 &0.48 &0.43 &0.25 &0.24 &0.44 &1.00 &0.39 &0.39 \\
    it $\times$ en &0.69 &0.27 &0.09 &0.42 &0.4 &0.42 &\textbf{0.94} &0.37 &0.37 \\
    en $\times$ it &0.22 &0.49 &0.32 &0.38 &0.16 &0.34 &0.76 &0.31 &0.31 \\
    it $\times$ it &\textbf{0.72} &0.36 &0.09 &0.51 &\textbf{0.35} &0.46 &0.82 &0.41 &0.41 \\
    en $\times$(en+it) &0.48 &0.47 &0.37 &0.04 &0.02 &0.37 &1.00 &0.27 &0.27 \\
    it $\times$(en+it) &0.43 &0.38 &0.16 &0.64 &0.06 &0.39 &\textbf{0.94} &0.34 &0.33 \\
    \bottomrule
    \end{tabular}
    
    }
    \caption{Llama3.1-8b model for moral foundation measurements on sampled MoralCovnITA dataset. The first language refers to the language of prompt while the languages after $\times$refer to the language(s) of fine-tuning datasets. $\varnothing$ denotes no fine-tuning. The best performing method is in \textbf{bold}. Column labels are consistent with Table \ref{tab:mt}.}
    \label{tab:italian}
\end{table}

Then we further qualitative assessed the performance on some moral foundation categories and revealed the cultural misalignment limitation in the finetuning process. Fine-tuning with English annotations sometimes produced negative effects. Performance on culturally sensitive foundations such as ``loyalty'' and ``sanctity'' declined significantly after fine-tuning with English and translated Italian data. This signals a concerning cultural misalignment in current LLMs --- fine-tuning with English-centric moral foundations may exacerbate inherent cultural biases. Our qualitative reading of some mislabeled Italian texts supports this finding. For example, Italian understanding of sanctity, influenced by Catholic values and historical context, differs significantly from American conceptions, resulting in poor performance after fine-tuning. Future research should further investigate fine-tuning techniques that mitigate cultural misalignment.

Moreover, we acknowledge the challenges involved in human annotation of moral foundations. Both CCV and MoralConvITA datasets have limitations, with annotator disagreements being common --- even the same annotator may label identical text differently at different times. Although training can reduce inconsistency, maintaining quality control remains difficult at scale. Both datasets relied on expert coders, which may introduce labeling biases. Additionally, MoralConvITA is subject to topic and text form biases, as it focuses on immigration during a specific period and consists of Twitter posts. In contrast, the CCV dataset covers a broader range of topics and primarily includes posts from news articles. Consequently, a general performance drop on the Italian dataset is expected. These challenges underscore the importance of creating high-quality benchmarking datasets for moral foundation measurement. Future research should consider employing demographically representative crowd workers to better capture the distribution of cultural values in local contexts \citep{hopp2021reflections}.

In conclusion, LLMs offer a promising approach for measuring moral foundations in non-English texts when limited local language annotations are available. With careful prompting and few-shot learning, LLMs deliver moderate performance compared to human annotation. It should be noted that fine-tuning with English annotations though may improve the performance on some moral foundation values, it can introduce additional biases to cultural-sensitive ones, therefore diminish the model performance.

\onecolumn
\begin{figure}[ht]
    \centering
    \includegraphics[width=0.7\pdfpagewidth]{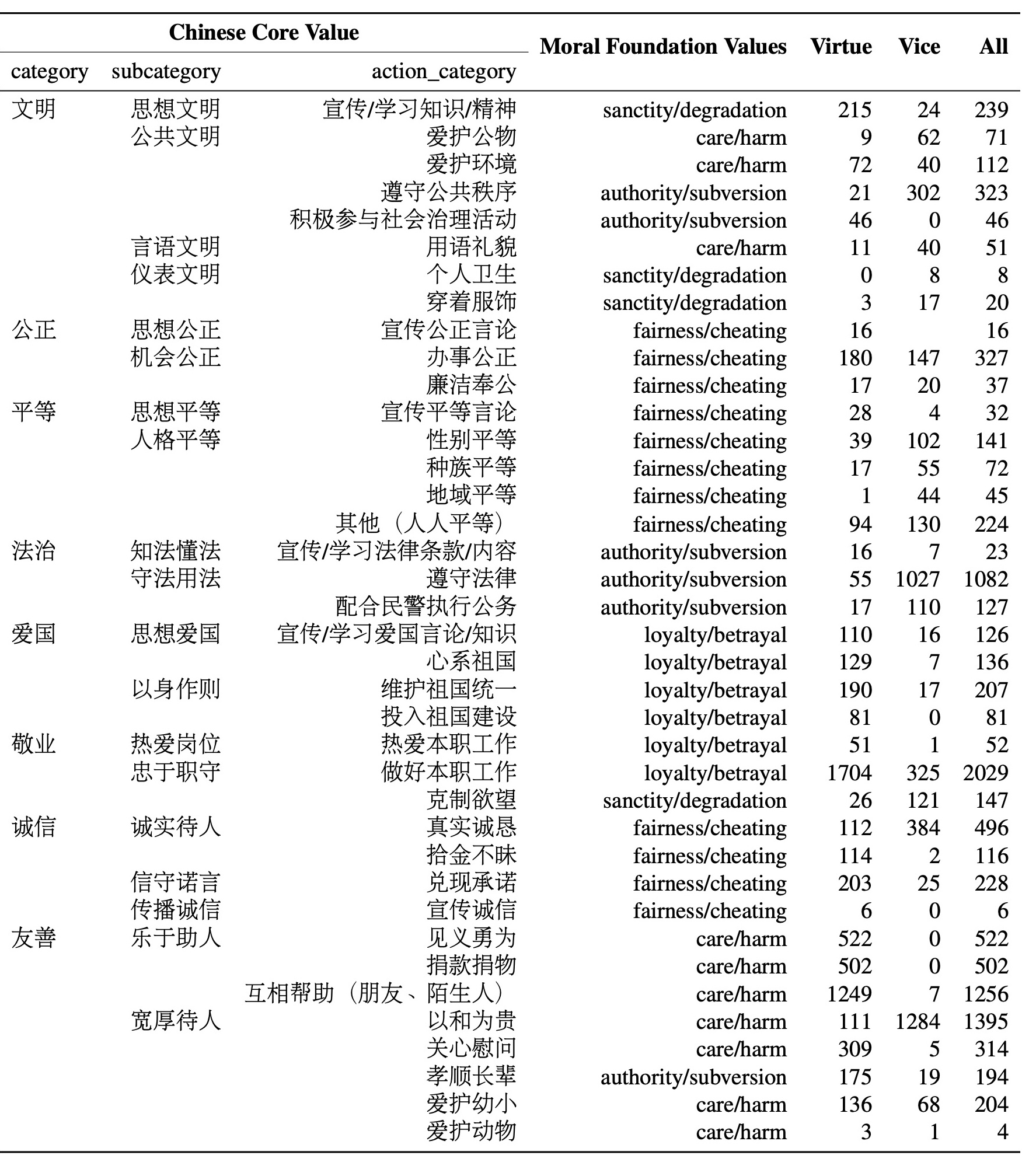} 
    \caption{Mapping Scheme of Chinese Core Value Dataset}
    \label{fig:cv_mapping}
\end{figure}

\begin{tcolorbox}[colback=gray!5!white, colframe=black!75, title=Prompt - English for CCV, fonttitle=\bfseries, boxrule=0.5pt]
\footnotesize
\begin{minipage}{\linewidth}
\begin{Verbatim}[breaklines=true, breakanywhere=true, fontsize=\footnotesize]
you are a native Chinese speaker and social science annotator, your task is to label the moral foundation values expressed in the given Chinese documents.

moral foundation values are the core values that underlie moral reasoning from the moral foundation theory. the five moral foundations are: care, fairness, loyalty, authority, and sanctity. And they refer to the following moral intuitions, each includes both vice/virtue pairs:
- care: related to our long evolution as mammals with attachment systems and an ability to feel (and dislike) the pain of others. It underlies the virtues of kindness, gentleness, and nurturance.
- fairness: related to the evolutionary process of reciprocal altruism. It underlies the virtues of justice and rights. 
- loyalty:  our long history as tribal creatures able to form shifting coalitions. It is active anytime people feel that its "one for all and all for one." It underlies the virtues of patriotism and self-sacrifice for the group. 
- authority: This foundation was shaped by our long primate history of hierarchical social interactions. It underlies virtues of leadership and followership, including deference to prestigious authority figures and respect for traditions.
- sanctity: This foundation was shaped by the psychology of disgust and contamination. It underlies notions of striving to live in an elevated, less carnal, more noble, and more natural way (often present in religious narratives). This foundation underlies the widespread idea that the body is a temple that can be desecrated by immoral activities and contaminants (an idea not unique to religious traditions). It underlies the virtues of self-discipline, self-improvement, naturalness, and spirituality. 

you should follow the given principles to label the moral foundation values in the give documents:
1. identify the moral foundation value only from the 5 given ones.
2. if the document expresses more than 1 foundation value, label all prominent values, but in total should be equal or less than  3 values.
3. provide a brief rationale for the each labelling, which should be less than 20 words. 
4. labels the value in english, 
5. rationales should be in the same lanaguage as the document
6. if the document does not express any of the 5 values, label it as 'none' and provide a brief rationale.
7. if the document can not be labelled into any of the 5 values, label it as 'unknown' and provide a brief rationale.
8. consider the Chinese cultural context of the document when labelling the values.

You MUST respond with a brief rationale within 15 words, and the labels. save in the dictionary format: {"rationale": "reasons to explain your decision", "labels": "you labels here"}

Here are the given documents for your task:
\end{Verbatim}
\end{minipage}
\end{tcolorbox}

\end{document}